\documentclass[conference]{IEEEtran} 

\usepackage{cite}
\usepackage{amsmath,amssymb,amsfonts}
\usepackage{algorithmic}
\usepackage{graphicx}
\usepackage{textcomp}

\usepackage[utf8]{inputenc}

\usepackage[dvipsnames]{xcolor}

\usepackage{caption}
\usepackage{subcaption}

\usepackage{tikz,pgfplots}
\usetikzlibrary{calc}
\usepgfplotslibrary{fillbetween}
\pgfplotsset{compat=1.10}

\usepackage{pgfplotstable}
\usetikzlibrary{positioning}
\usetikzlibrary{pgfplots.groupplots}
\usetikzlibrary{arrows}
\usetikzlibrary{matrix}
\usepackage{multirow} 
\usepackage{amsthm}

\newcommand{\gennorm}{\mathsf{GenNorm}}
\newcommand{\norm}{\mathsf{Norm}}
\usepackage{hyperref}
\hypersetup{
    colorlinks=true,
    linkcolor=black,
    filecolor=magenta,      
    urlcolor=cyan,
    pdfpagemode=FullScreen,
    }
    
{
\theoremstyle{plain}
\newtheorem*{assumption*}{Assumption}
}

\usepackage{siunitx}
\usepackage{mathcomSTEv4}

 \newtheorem{remark}{Remark}
 
\begin{document}

\title{DNN gradient lossless compression: \\
Can GenNorm be the answer?}

\author{Zhong-Jing Chen\\
NYCU, Taiwan\\
zhongjing.ee10@nycu.edu.tw
\and
Eduin E. Hernandez\\
NYCU, Taiwan \\
eduin.ee08@nycu.edu.tw
\and
Yu-Chih Huang\\
NYCU, Taiwan\\
jerryhuang@nycu.edu.tw
\and
Stefano Rini\\
NYCU, Taiwan \\
stefano.rini@nycu.edu.tw
}

\maketitle
\begin{abstract}

In this paper, the problem of optimal gradient lossless compression in Deep Neural Network (DNN) training is considered. 
Gradient compression is relevant in many distributed DNN training scenarios, including the recently popular federated learning (FL) scenario in which each remote users are connected to the parameter server (PS)  through a noiseless but rate limited channel.
In distributed DNN training, if the underlying gradient distribution is available, classical lossless compression approaches can be used to reduce the number of bits required for communicating the gradient entries.
Mean field analysis has suggested that gradient updates can be considered as independent random variables, while Laplace approximation can be used to argue that gradient has a distribution approximating the  normal ($\norm$) distribution in some regimes. 
In this paper we argue that, for some networks  of practical interest, the gradient entries can be well  modelled as having a generalized  normal ($\gennorm$) distribution.
We provide numerical evaluations to validate that the hypothesis $\gennorm$ modelling provides a more accurate prediction of the DNN  gradient tail distribution. 
%
%
Additionally, this modeling choice provides concrete improvement in terms of lossless compression of the gradients when applying classical fix-to-variable lossless coding  algorithms, such as Huffman coding, to the quantized gradient updates. 
This latter results indeed provides an effective compression strategy with low memory and computational complexity that has great practical relevance in distributed DNN training scenarios. 
\end{abstract}
\begin{IEEEkeywords}
DNN training; Distributed optimization; Lossless compression; Generalized normal distribution.
\end{IEEEkeywords}

\section{Introduction}

The digitalization of physical devices,  system infrastructure, and data services, as embodied by the IoT paradigm, has  enabled the collection of  large-scale databases which can be used for a trove of machine learning (ML) tasks, ranging from autonomous driving to health-care services and smart energy management. 
In this paradigm, data centralization is no longer a feasible 
%
and thus distributed ML is being hailed as the next milestone in large-scale data computing.

Among the various distributed ML architectures, federated learning (FL) has received particular attention: FL consists of a central model which is trained   locally at the remote clients by applying stochastic gradient descent (SGD) over a local dataset.
The local gradient are then communicated to the central parameter server (PS) for aggregation into a global model.
For this model, one useful concept is that of \emph{communication overhead} \cite{shlezinger2020communication}, that is the number of bits-per-iteration that are required by a decentralize  training scheme to attain a certain convergence guarantee for the central model as compared to the centralized training case.
The concept of communication overhead naturally points to the model in which the communication between the  remote user and the PS is subject to a total
constraint in the number of bits exchanged throughout training. 
%
In this paper we investigate this setting and focus on determining the relevant assumptions under which an efficient compression schemes can be used for minimizing the communication overhead for the DNN training scenario.

%
%
\smallskip
\noindent
\underline{\bf Relevant Literature:}
%
In the following, we shall discuss communication aspects of FL and distributed training relevant to the development of the paper. 
Various approaches have been proposed in the literature to improve  communication efficiency in FL. 
The dimensionality-reduction schemes put forth in the literature  mainly fall into two categories: gradient sparsification, 
\cite{Shalev-Shwartz2010FL_CE,
Alistarh2018Spars_FL, Amir_FL}, 
and gradient quantization  \cite{seide2014onebitSGD,Konecny2016Fl_CE, gandikota2019vqsgd}.
%
%
Most of the proposed dimensionality-reduction techniques are applied to each of the gradient dimension separately, such as 
quantized SGD (QSGD) and its stochastic versions.
%
%
Dimensionality-reduction can also be performed on the whole gradient vector as suggested in \cite{gandikota2019vqsgd} through an algorithm referred to as vector QSGD (VQSGD).
%
From a more implementation-oriented perspective, \cite{sun2019hybrid} studies the effect of gradient quantization in the $8$-bit floating-point ($8$-fp) representation as \emph{sign-exponent-mantissa}, which is commonly adopted in numerical implementations of DNN training,


\smallskip
\noindent
\underline{\bf Contributions:}
%
In the following, we focus on the design of lossless compression schemes for the compression of DNN gradient after $8$-fp quantization. 
In particular,  we aim at providing a good statistical model for DNN gradients training, that  can be effectively used to design both quantization and compression schemes in decentralized training scenarios.
%
%
Our main contributions are summarized as follows:
%
 
     
 \noindent
 \underline{\emph{GenNorm modelling:}} 
 To the best of our knowledge, a good statistical model for modelling gradients in DNN training is currently lacking.
 We argue that one can effectively model such gradients as i.i.d. random variables having a generalized normal distribution, which we refer to as $\gennorm$. 
 We use statistical methods to validate the $\gennorm$ assumption for three DNN architectures in the image classification task, namely DenseNet\cite{huang2017densely},  ResNet\cite{he2016identity}, and  NASNet\cite{zoph2018learning},
 across both layers and training epochs.
 We also  argue that (i) the gradient distribution approaches the normal ($\norm$) distribution as the depth of the network increases and as the epoch number increases, additionally (ii) we contend that the kurtosis of the gradient distribution provides a useful measure of the concentration of gradient around zero.
     
\noindent
 \underline{\emph{GenNorm gradient compression performance and its $8$-fp}}
 \underline{\emph{quantization performance:}}
     We investigate the communication overhead that can be attained through lossless compression of DNN gradients after $8$-fp quantization. 
To argue for the effectiveness of our assumption and the DNN gradient distribution, we compare
 %
 %
 the case in which the gradients compressed using the $\gennorm$ assumption versus (i) the standard $\norm$ assumption, and (ii) using an universal compression in the form of LZ78 \cite{ziv1978compression}. 
 Our results show that the required communication overhead with the $\gennorm$ model is much less than that with the standard $\norm$ model in the upper and middle layers and the performance of both distributions gradually become the same towards the lower layers. Moreover, they both have significant gains over LZ78 in all the layers.
 %
%
%
%
%
 Further investigations on the theoretical foundations of the $\gennorm$ assumptions are left for future work; here we shall only focus on the  numerical evaluations of this assumption. 

\noindent
{\bf Notation.}
Lowercase boldface letters (e.g., $\zv$) are used for column vectors,  
uppercase letters for random variables (e.g. $X$), and calligraphic uppercase for  sets (e.g. $\Acal$) .
We also adopt the shorthands $[m:n] \triangleq \{m, \ldots, n\}$
and  $[n] \triangleq \{1, \ldots, n\}$. 
Additionally  $\Gamma(.)$ denotes the Gamma function, $\norm$ the normal distribution, and $\gennorm$ the generalized normal distribution. 
Finally, $\Fbb_2$ is the binary field.
%


\section{Related Results}
\label{sec:Related Results}

\subsection{Federated Learning and Federated Averaging 
}\label{subsec:FL}
%
%
%
The FL model consists of $U$ remote users communicating their local gradient to the PS over $T$ iterations with the aim of  training a global ML model capable of optimizing a given loss function
$\ell\lb\ldotp\rb$ obtained as the average of the local loss functions $\ell^{(u)}$ at each of the users $u$, as evaluated on the local dataset
%
\ea{
\Dv^{(u)} = \left\{\lb \dv_{k}^{\lb u\rb},v_k^{\lb u\rb}\rb\right\}_{k\in\left[\left|\Dc^{(u)}\right|\right]}.
}
The local dataset $\Dv^{(u)}$   includes
$\left|\Dv^{(u)}\right|$
pairs, each comprising of a data point 
$\dv_{k}^{\lb u\rb}$
and the label $v_{k}^{(u)}$.
The remote users collaborate with the PS during $T$ iterations in order to minimize the loss function and find the solution $\wv^\ast$  defined as
\ea{ \label{eq:optim_value}
\wv^{\ast} 
& =\arg\min_{\wv}\frac{1}{\left|\Dcal\right|}
=
\sum_{u\in[U]}\left|\Dcal^{(u)}\right|\ell^{(u)}\left(\wv_{t};\dv^{(u)}_{k},v^{(u)}_{k}\right),
}
where $\ell^{(u)}\lb\ldotp\rb$ is the local loss function at remote user $u$. 
%
%

A common approach for numerically determining the optimal value in \eqref{eq:optim_value} is through iterative application of synchronous (stochastic) gradient descent (SGD).
In the SGD algorithm,  the model parameter $\wv$ is updated at each iteration $t\in[T]$ in the negative direction of the gradient vector multiplied by an iteration-dependent step size $\gamma_t$ called the \emph{learning rate}, as in \eqref{eq:model_update}.
In the federated setting, SGD can be implemented by having each remote user communicate the local gradients to the PS. 
The PS aggregates the local gradients so as to obtain a global gradient which is employed in the global model update.
The resulting algorithms is customarily referred on as \emph{federated averaging} \cite{Konecny2016Fl_CE}.
%
%
%
Note that, in federated averaging, the local gradient $\gtv_{t}^{(u)}$ is computed as 
%
\ea{\label{eq:stochastic_local_gradient}
\gtv_{t}^{(u)}
& =\dfrac{1}{\left|\Dcal^{(u)}\right|}\sum_{k\in\left[\left|\Dcal^{(u)}\right|\right]} \nabla\ell^{(u)}\left(\wv_{t};\dv^{(u)}_{k},v^{(u)}_{k}\right).
}
%
The global gradient of the loss function $\ell\lb\ldotp\rb$ at iteration $t$ is computed at the PS by aggregating the received local gradients according to a distributed mean estimation (DME) as
\ea{\label{global_SG}
\gov_{t}=\f{1}{U}\sum_{u\in [U]}\gtv_{t}^{(u)}.
}
Next, the global or final model at iteration $t+1$ is updated as
\ea{\label{eq:model_update}
\wv_{t+1}=\wv_{t}-\gamma_{t}\gov_{t}.
}
The convergence of federated averaging can be shown under various assumptions on the loss function \cite{konevcny2016federated}.
%

\subsection{Mean Field theory}
In recent years, the mean field theory has been applied to the study of DNN and has achieved a great deal of success.
%
%
Consider a simple DNN with  two layers
minimizing the square loss over an i.i.d. dataset, \cite{mei2018mean} shows that training through SGD is well-approximated  by continuous dynamics expressed through certain non-linear partial differential equation.
In  \cite{mei2019mean} this analysis is further extended to study the empirical distribution of the neurons after $k$ SGD steps.
It is shown that, under the assumption of i.i.d. initialization of the weights, the weight distribution evolves according to a particular stochastic differential equation.
Note that the  analysis of SGD dynamics has been developed that connects naturally to
the theory of universal approximation \cite{sirignano2020mean}.




\section{System Model}
\label{sec:System Model}
In many distributed training scenarios of practical relevance, such as FL paradigm, 
%
%
the communication from the server to the remote users is  unconstrained, as the PS is generally not limited in power or connectivity.
Accordingly, the main bottleneck  is the the up-link communication \cite{konevcny2016federated,li2019federated} 
	i.e., the updates transferred from the users to the centralized servers.
For this reason,  we  study distributed training scenario in which a remote user wishes to communication its local stochastic gradient in \eqref{eq:stochastic_local_gradient} to PS. 
To measure the transmission efficiency in the decentralized model training scenario, we  introduce a measure of communication complexity as in the next section.


\subsection{Rate-limited distributed DNN training}



Consider the distributed DNN training scenario in which the
communication between each user and the PS take place over a noiseless channel with finite capacity. 
To meet the finite capacity constraint, the local gradient $\gtv_{t}^{(u)}$ is first quantized via a quantizer $Q:\mathbb{R}\rightarrow \mathcal{X}$ to form the representative $\hat{\gv}_t^{(u)}=Q(\gtv_{t}^{(u)})$, where $\mathcal{X}$ is the collection of representatives (i.e., quantization levels).

After that, we employ data compression $h:\mathcal{X}\rightarrow \mathbb{F}_2^{*}$ to form $\mathbf{b}_t^{(u)}=h(\hat{\gv}_t^{(u)})$, which removes the redundancy inherent in the local gradients for reducing the amount of data required to be transmitted. 
Note that we allow $h$ to be a variable-length coding scheme; hence, the range is $\mathbb{F}_2^{*}$.  

Let us assume that  the local gradient is distributed i.i.d. according to $\mathbb{P}_{\tilde{G}_t}$. 
Also, let $r_t^{(u)}$ be the length of $\mathbf{b}_t^{(u)}$. We define the expected length of $u\in[U]$ at $t\in[T]$ as
\ea{
R_t^{(u)} = \mathbb{E}_{Q,h} \left[r_t^{(u)}\right],
}
where the expectation is taken w.r.t. $\mathbb{P}_{\tilde{G}_t}$. 
We are now ready to define the {\em communications overhead} of a certain pair of $(Q,h)$ as the  \underline{sum expected lengths} conveyed over the up-link channel  over the training, that is 
	\begin{equation}
	\label{eqn:Overhead}
	R = \sum_{t \in [T]} \sum_{u \in [U]} R_t^{(u)}.
	\end{equation}
In this paper, we consider the \emph{lossless compression} scenarios in which the PS is interested in the exact reconstruction of the quantized gradients $\hat{\gv}_t^{(u)}$ from $\mathbf{b}_t^{(u)}$.
When lossless compression is considered, classical results in lossless source coding can be applied for gradient compression. 
When the underlying distribution $\mathbb{P}_{\tilde{G}_t}$ is unknown, one can employ Lempel-Ziv coding, which is asymptotically optimal in terms of the expected length. 
However, the performance of such an universal source coding scheme is not acceptable in the short to medium source length regime. 
In contrast, in the presence of knowledge about $\mathbb{P}_{\tilde{G}_t}$, optimal lossless compression can be easily achieved by Huffman coding \cite{cover_book}. 
This naturally raises the problem of statistical modeling of gradient distribution, which is the main focus of this paper.

\begin{remark} 
Throughout the paper, we shall not investigate the lossy gradient compression case. 
This follows from the fact that a precise understanding of the effect of the distortion criteria used for compression on the learning performance is unclear. 
For instance, top-k sparsification  \cite{shi2019understanding} suggests that an appropriate choice of  distortion should take into account the gradient magnitude.
This is in contrast with the compression error introduced by the classic MSE criteria which is commonly used in practical lossless compression algorithms.
%

\end{remark}


\section{Proposed Approach} \label{sec:Proposed Approach}
Let us begin by clarifying the simulations setting used in the remainder of the section. 
After that, we present our main contributions, which contain a set of simulations that allow us to argue that $\gennorm$ is a fair model for gradient distribution. Our evidence includes comparisons based on histogram, Wasserstein distance of order 2, and compression rates with Huffman codes. 

 \subsection{DNN training setting}
\label{subsec:DNN train setting}
In this paper, we consider the training for the CIFAR-10 dataset classification task using the following three architectures: (i) DenseNet121, (ii) ResNet50V2, and (iii) NASNetMobile. 
For each architecture, the training is performed using SGD optimizer with a constant $\gamma_{t}=0.01$ learning rate in \eqref{eq:model_update}.
The rest of the configurations of the parameters and hyperparameters used for the training are specified in Tab. \ref{tab:DNN parameters}.

\begin{table}
	\footnotesize
	\centering
	\vspace{0.04in}\caption{Parameters and hyperparameters used for the training of the DNN models.}
	\label{tab:DNN parameters}
	\begin{tabular}{|c|c|}
		\hline
		Dataset & CIFAR-10 \\ \hline
        Training Samples & \num{50000} \\ \hline
        Test Samples & \num{10000} \\\hline
        Optimizer & SGD \\ \hline
        Learning Rate & \num{0.01} \\ \hline
        Momentum & 0 \\ \hline
        Loss & Categorical Cross Entropy \\ \hline
        Epochs & 100 \\ \hline
        Mini-Batch Sizes & 64 \\ \hline
	\end{tabular}
\end{table}

During each batch-iterations, the gradients of the trainable parameters are accumulated on a temporal memory on a per layer basis with the intention on averaging them along the epoch.
At the end of the epoch, the gradients are saved and the temporal memory is freed.
This process is repeated until the last epoch for the gradient analysis provided in the next subsections.\footnote{The code for the gradient modeling and analysis is available at  \url{https://github.com/Chen-Zhong-Jing/Save_Model_Gradient}}
As these are very deep networks as specified in Tab. \ref{tab:total network parameters}, we will limit the scope to three layers in each of the architectures: one 2-dimensional convolution layer located in the upper, middle, and lower sections of the networks.
Tab. \ref{tab:network parameters} details the number of trainable weight parameters for these chosen layers.
\begin{table}
	\footnotesize
    \caption{Total number of layers, weight parameters, and trainable weight parameters belonging to each architecture.} \label{tab:total network parameters}
    \centering
    \begin{tabular}{|c|c|c|c|}
    \hline{Architectures}
    & Layers & Total Params & Train Params \\
    \hline{DenseNet121}
    & 121 & \num{7047754} & \num{6964106} \\
    \hline{ResNet50V2}
    & 50 & \num{23585290} & \num{23539850} \\
    \hline{NASNetMobile}
    & - & \num{4280286} & \num{4243548} \\
    \hline
    \end{tabular}
\end{table}
\begin{table}
    \footnotesize
    \caption{Number of trainable weight parameters for the chosen layers of each architecture.} 
    \label{tab:network parameters}
    \centering
    \begin{tabular}{|c|c|c|c|}
    \hline{Architectures}
    & Upper & Middle & Lower\\
    \hline{ResNet50V2}
    & \num{4096} & \num{32768} & \num{524288} \\
    \hline{DenseNet121}
    & \num{9408} & \num{16384} & \num{65536} \\
    \hline{NASNetMobile}
    & \num{3872} & \num{30976} & \num{185856} \\
    \hline
    \end{tabular}
\end{table}

\subsection{Gradient Quantization}
For the quantizer $Q$, we adopt the $8$-bit sign-exponent-mantissa with $[1,5,2]$ to quantize the gradients. The $8$-bit $[1,5,2]$ format forms the range $[2^{-16},2^{15}]$ of numbers, which are used to establish the bin edges. The gradients are quantized to the centers of the bins they locate in.


\subsection{GenNorm modeling}
\label{sec:GenNorm modeling}

First, we wish to argue that the gradient distribution $\mathbb{P}_{G_t}$ can be modelled as an i.i.d. $\gennorm$ distribution, i.e., they have the pdf 
\ea{
\gennorm(x,\mu, \al, \be) 
= \f {\be}{2\al \Gamma(1/\be)} \exp \lcb 
-  \lb \f  { \labs x-\mu\rabs} {\al} \rb^{\be}\rcb,  
\label{eq:gennorm}
}
where $(\mu, \al, \be)$ are the location, scale, and shape parameters, respectively. Some important parameters for $\gennorm$ includes the mean, variance, and kurtosis that have the following expressions:
\ea{
\mathrm{Mean}=\mu, \quad
\mathrm{Var} = \frac{\alpha^2\Gamma(3/\beta)}{\Gamma(1/\beta)}, \quad
\mathrm{Kurt} = \frac{\Gamma(5/\beta)\Gamma(1/\beta)}{\Gamma(3/\beta)^2}.
\label{eq:kurtosis}
}

$\gennorm$ is a family of distributions that subsumes Laplace ($\be=1$) and Normal $(\be=2)$ distributions. 
When the shape parameter $\be < 2$, the distribution is leptokurtic and has fatter tail than the normal distribution.

\begin{assumption*}{\bf GenNorm DNN gradients:}
For each layer and each epoch, the DNN gradients are distributed according to the $\gennorm$ distribution in \eqref{eq:gennorm}.
\end{assumption*}
We refer to the above assumption as the \emph{GenNorm} assumption.
In the remainder of the section, we shall motivate the $\gennorm$ assumption from a statistical perspective. 
Successively, we shall motivate this assumption from a practical perspective by showing that it offers substantial advantages for the setting in Sec.  \ref{sec:System Model}.

\medskip

\begin{figure}
    \centering
    \begin{tikzpicture}
    \definecolor{mycolor1}{rgb}{0.00000,0.44706,0.74118}%
    \definecolor{mycolor2}{rgb}{0.63529,0.07843,0.18431}%
    \definecolor{mycolor3}{rgb}{0.00000,0.49804,0.00000}%
    \begin{groupplot}[
        group style={
            group name=my plots,
            group size=2 by 3,
            xlabels at=edge bottom,
            xticklabels at=edge bottom,
            vertical sep=0pt,
            horizontal sep=5pt
        },
        height=5cm,
        width=5.5cm,
        xmax=0.006,
        xmin=-0.006,
        ymax=1100,
        ymin=0,
        xlabel={x},
        xtick={-0.005, -0.0025, 0, 0.0025, 0.005},
        xmajorgrids,
        ymajorgrids
    ]
    \nextgroupplot[title=ResNet50V2,
    yticklabels=\empty,
    ylabel={Epoch 2}]
        \coordinate (top) at (axis cs:0,\pgfkeysvalueof{/pgfplots/ymin}) -- (axis cs:0,\pgfkeysvalueof{/pgfplots/ymax});
        \addplot+[ybar interval,mark=no,fill=blue!120,draw=blue, opacity=0.5]
                [restrict x to domain=-0.005:0.005]
                table[x index=0, y index=1]{./Data/ResNet50V2_top_layer_e2_hist_without_quantization.txt};
        \addplot [fill=red!120,draw=red, opacity=0.4]
                [restrict x to domain=-0.005:0.005]
                table[x index=0, y index=1]{./Data/ResNet50V2_top_layer_e2_pdf_without_quantization.txt};
        \addplot [fill=green!120,draw=green, opacity=0.35]
                [restrict x to domain=-0.005:0.005]
                table[x index=0, y index=2]{./Data/ResNet50V2_top_layer_e2_pdf_without_quantization.txt};
    \nextgroupplot[title=NASNetMobile,yticklabels=\empty]
        \coordinate (top) at (axis cs:0,\pgfkeysvalueof{/pgfplots/ymin}) -- (axis cs:0,\pgfkeysvalueof{/pgfplots/ymax});
        \addplot+[ybar interval,mark=no,fill=blue!120,draw=blue, opacity=0.5]
                [restrict x to domain=-0.005:0.005]
                table[x index=0, y index=1]{./Data/NASNetMobile_top_layer_e2_hist_without_quantization.txt};
        \addplot [fill=red!120,draw=red, opacity=0.4]
                [restrict x to domain=-0.005:0.005]
                table[x index=0, y index=1]{./Data/NASNetMobile_top_layer_e2_pdf_without_quantization.txt};
        \addplot [fill=green!120,draw=green, opacity=0.35]
                [restrict x to domain=-0.005:0.005]
                table[x index=0, y index=2]{./Data/NASNetMobile_top_layer_e2_pdf_without_quantization.txt};
    \nextgroupplot[yticklabels=\empty,
    ylabel={Epoch 50}]
        \addplot+[ybar interval,mark=no,fill=blue!120,draw=blue, opacity=0.5]
                [restrict x to domain=-0.005:0.005]
                table[x index=0, y index=1]{./Data/ResNet50V2_top_layer_e50_hist_without_quantization.txt};
        \addplot [fill=red!120,draw=red, opacity=0.4]
                [restrict x to domain=-0.005:0.005]
                table[x index=0, y index=1]{./Data/ResNet50V2_top_layer_e50_pdf_without_quantization.txt};
        \addplot [fill=green!120,draw=green, opacity=0.35]
                [restrict x to domain=-0.005:0.005]
                table[x index=0, y index=2]{./Data/ResNet50V2_top_layer_e50_pdf_without_quantization.txt};
    \nextgroupplot[yticklabels=\empty]
        \addplot+[ybar interval,mark=no,fill=blue!120,draw=blue, opacity=0.5]
                [restrict x to domain=-0.005:0.005]
                table[x index=0, y index=1]{./Data/NASNetMobile_top_layer_e50_hist_without_quantization.txt};
        \addplot [fill=red!120,draw=red, opacity=0.4]
                [restrict x to domain=-0.005:0.005]
                table[x index=0, y index=1]{./Data/NASNetMobile_top_layer_e50_pdf_without_quantization.txt};
        \addplot [fill=green!120,draw=green, opacity=0.35]
                [restrict x to domain=-0.005:0.005]
                table[x index=0, y index=2]{./Data/NASNetMobile_top_layer_e50_pdf_without_quantization.txt};
    \nextgroupplot[yticklabels=\empty,
    ylabel={Epoch 100}]
        \addplot+[ybar interval,mark=no,fill=blue!120,draw=blue, opacity=0.5]
                [restrict x to domain=-0.005:0.005]
                table[x index=0, y index=1]{./Data/ResNet50V2_top_layer_e100_hist_without_quantization.txt};
        \addplot [fill=red!120,draw=red, opacity=0.4]
                [restrict x to domain=-0.005:0.005]
                table[x index=0, y index=1]{./Data/ResNet50V2_top_layer_e100_pdf_without_quantization.txt};
        \addplot [fill=green!120,draw=green, opacity=0.35]
                [restrict x to domain=-0.005:0.005]
                table[x index=0, y index=2]{./Data/ResNet50V2_top_layer_e100_pdf_without_quantization.txt};
    \nextgroupplot[yticklabels=\empty]
        \addplot+[ybar interval,mark=no,fill=blue!120,draw=blue, opacity=0.5]
                [restrict x to domain=-0.005:0.005]
                table[x index=0, y index=1]{./Data/NASNetMobile_top_layer_e100_hist_without_quantization.txt};
        \addplot [fill=red!120,draw=red, opacity=0.4]
                [restrict x to domain=-0.005:0.005]
                table[x index=0, y index=1]{./Data/NASNetMobile_top_layer_e100_pdf_without_quantization.txt};
        \addplot [fill=green!120,draw=green, opacity=0.35]
                [restrict x to domain=-0.005:0.005]
                table[x index=0, y index=2]{./Data/NASNetMobile_top_layer_e100_pdf_without_quantization.txt};
    \end{groupplot}
\end{tikzpicture}
    \caption{Histogram of gradient (blue) with PDF of $\gennorm$ (red) and $\norm$ (green) of top layer for epoch 2, 50, and 100 in two different network.}
    \label{fig:histogram}
\end{figure}
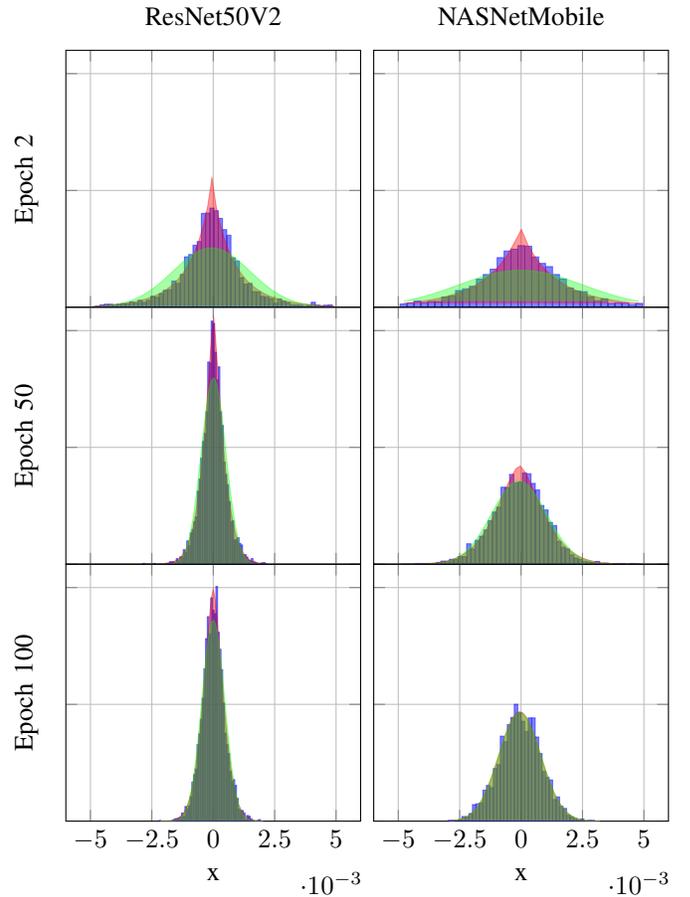

\noindent
{\bf Stochastic validation:}
Let us begin by visually inspecting the gradient histogram for the networks in Sec.  \ref{subsec:DNN train setting}, as depicted in 
Fig. \ref{fig:histogram}.
In this figure, we plot (i) the sample distribution, (ii) the $\gennorm$ fitting, and (iii) the $\norm$ fitting for ResNet50V2 and NASNetMobile across three epoch: $2$, $50$, and $100$.
We observe that in the earlier epochs, the gradient histogram is closer to the $\gennorm$ distribution in that the sample distribution is (i) more concentrated in zero, and  (ii) it contains heavier tails than the $\norm$ distribution. 
As the training continues, the  variance of the gradient distribution gradually reduces and approaches the $\norm$ distribution.
For instance, the gradients from ResNet50V2 seems to converge to the $\norm$ distribution slower than NASNetMobile.

\begin{figure}
    \centering
    \begin{tikzpicture}[scale = 0.8]
    \definecolor{mycolor1}{rgb}{0.00000,0.44706,0.74118}%
    \definecolor{mycolor2}{rgb}{0.63529,0.07843,0.18431}%
    \definecolor{mycolor3}{rgb}{0.00000,0.49804,0.00000}%
    \begin{axis}[
    ymin=0.05,
    xmin=0,
    xmax=100,
    xlabel={Epoch},
    ylabel={Wasserstein Distance},
    grid=both]
        \addplot [draw=mycolor1, line width=1.5pt, mark=diamond, mark options={solid, mycolor1}, mark repeat={10}, smooth]
            table[x index=0, y index=3]{./Data/DenseNet121_mid_end_w.txt};
            \addlegendentry{GenNorm}
        \addplot [draw=mycolor2, line width=1.5pt, mark=square, mark options={solid, mycolor2}, mark repeat={10}, smooth]
            table[x index=0, y index=4]{./Data/DenseNet121_mid_end_w.txt};
            \addlegendentry{Norm}
    \end{axis}
\end{tikzpicture}
    \vspace{-0.3cm}
    \caption{1D Wasserstein distance of the DenseNet121 lower layer.}
    \label{fig:gradient_wasserstein_distance_2}
\end{figure}
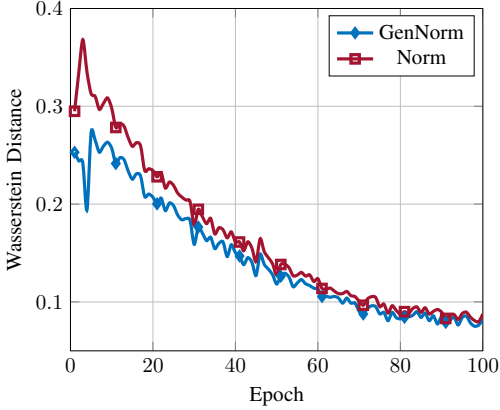

Fig. \ref{fig:histogram} only provides a qualitative depiction of the $\gennorm$ assumption. 
A quantitative depiction is provided in Fig. \ref{fig:gradient_wasserstein_distance_2}: here we plot the 1D $W_2$ Wasserstein distance, defined as 
%
\ea{
W_2(X,Y)=\left(\int_{0}^{1}|F^{-1}_{X}(z)-F^{-1}_{Y}(z)|dz\right)^{1/2} 
}
between the $\gennorm$ distribution and the gradient samples versus the $\norm$ distribution and the gradient samples for the lower layer DenseNet121 as a function of the epoch number.
We again notice that the $\gennorm$ provides a closer fitting with 
gradient samples than the $\norm$. 
Additionally, we notice the relative distance between $\gennorm$ and $\norm$ decreases with the epoch number, again suggesting that for large enough epoch number the gradient distribution tends towards the $\norm$.

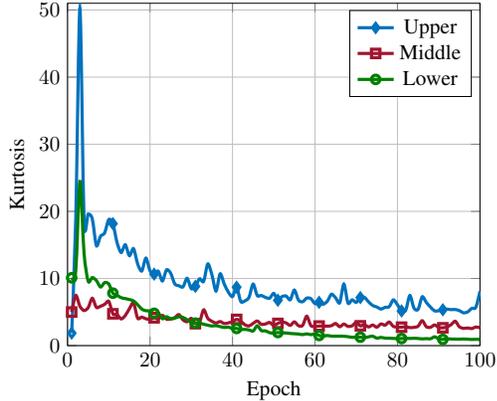
\begin{figure}
    \centering
    \begin{tikzpicture}[scale = 0.8]
    \definecolor{mycolor1}{rgb}{0.00000,0.44706,0.74118}%
    \definecolor{mycolor2}{rgb}{0.63529,0.07843,0.18431}%
    \definecolor{mycolor3}{rgb}{0.00000,0.49804,0.00000}%
    \begin{axis}[
    ymin=0,
    ymax=51,
    xmin=0,
    xmax=100,
    xlabel={Epoch},
    ylabel={Kurtosis},
    grid=both]
        \addplot [draw=mycolor1, line width=1.5pt, mark=diamond, mark options={solid, mycolor1}, mark repeat={10}, smooth]
            table[x index=0, y index=1]{./Data/DenseNet121_kur.txt};
            \addlegendentry{Upper}
        \addplot [draw=mycolor2, line width=1.5pt, mark=square, mark options={solid, mycolor2}, mark repeat={10}, smooth]
            table[x index=0, y index=2]{./Data/DenseNet121_kur.txt};
            \addlegendentry{Middle}
        \addplot [draw=mycolor3, line width=1.5pt, mark=o, mark options={solid, mycolor3}, mark repeat={10}, smooth]
            table[x index=0, y index=3]{./Data/DenseNet121_kur.txt};
            \addlegendentry{Lower}
    \end{axis}
\end{tikzpicture}
    \vspace{-0.3cm}
    \caption{Kurtosis of the gradients of the Upper, Middle, and Lower convolution layer in the DenseNet121.}
    \label{fig:densenet_kurtosis}
\end{figure}

\begin{figure}
    \centering
    \begin{tikzpicture}[scale = 0.8]
    \definecolor{mycolor1}{rgb}{0.00000,0.44706,0.74118}%
    \definecolor{mycolor2}{rgb}{0.63529,0.07843,0.18431}%
    \definecolor{mycolor3}{rgb}{0.00000,0.49804,0.00000}%
    \begin{axis}[
    ymin=0,
    ymax=58,
    xmin=0,
    xmax=100,
    xlabel={Epoch},
    ylabel={Kurtosis},
    grid=both]
        \addplot [draw=mycolor1, line width=1.5pt, mark=diamond, mark options={solid, mycolor1}, mark repeat={10}, smooth]
            table[x index=0, y index=1]{./Data/NASNetMobile_kurtosis.txt};
            \addlegendentry{Upper}
        \addplot [draw=mycolor2, line width=1.5pt, mark=square, mark options={solid, mycolor2}, mark repeat={10}, smooth]
            table[x index=0, y index=2]{./Data/NASNetMobile_kurtosis.txt};
            \addlegendentry{Middle}
        \addplot [draw=mycolor3, line width=1.5pt, mark=o, mark options={solid, mycolor3}, mark repeat={10}, smooth]
            table[x index=0, y index=3]{./Data/NASNetMobile_kurtosis.txt};
            \addlegendentry{Lower}
    \end{axis}
\end{tikzpicture}
    \vspace{-0.3cm}
    \caption{NASNetMobile: Kurtosis of each convolution layer.}
    \label{fig:nasnet_kurtosis}
\end{figure}
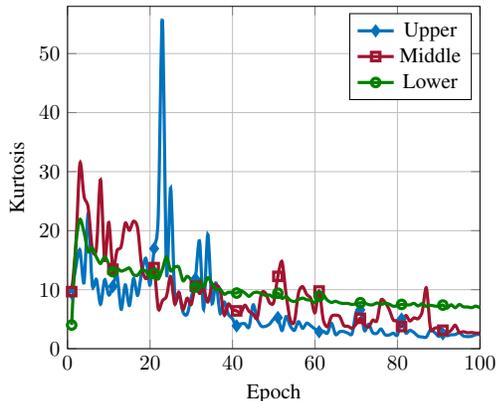

\smallskip
\noindent
{\bf Distribution parameters:}
The mean and variance of the sample gradient distribution is provided in Table. \ref{tab:network mean and variance}, together with the respective  confidence interval. 
Another important aspects of the $\gennorm$ is that it highlights the role of the kurtosis in describing the behaviour of the gradients, as in \eqref{eq:kurtosis}, the kurtosis depends only on the parameter $\be$.
From Figs. \ref{fig:densenet_kurtosis} and \ref{fig:nasnet_kurtosis}, we again observe the $\gennorm$ modeling tending towards the $\norm$ with further epochs:
The excess kurtosis evolves from positive to near zero.
%

\begin{table*}[t]
    \footnotesize
    \vspace{0.04in}\caption{The mean and variance of NASNetMobile's gradient at different epochs} 
    \label{tab:network mean and variance}
    \centering
    \begin{tabular}{|c|c|c|c|c|}
    \hline
    
    \hline{Layers}
    & & Epoch 2 & Epoch 50 & Epoch 100\\
    \hline
    \multirow{2}{*}{Upper}
    & mean 
    & \num{-1.72e-05} $\pm$ \num{1.12e-07}
    & \num{-8.38e-05} $\pm$ \num{1.90e-08}
    & \num{-5.53e-05} $\pm$ \num{2.13e-08}
    \\ \cline{2-5}
    & variance 
    & \num{9.58e-05}$\pm$ \num{8.03e-08}
    & \num{1.86e-05}$\pm$ \num{3.82e-10}
    & \num{1.07e-05}$\pm$ \num{1.82e-11}
    \\ \hline
     \multirow{2}{*}{Middle}
    & mean 
    & \num{6.66e-06}$\pm$ \num{9.53e-09}
    & \num{1.10e-05}$\pm$ \num{8.57e-09}
    & \num{3.68e-06}$\pm$ \num{1.92e-09} 
    \\ \cline{2-5}
    & variance 
    & \num{4.96e-06}$\pm$ \num{1.19e-10}
    & \num{3.56e-06}$\pm$ \num{7.66e-11}
    & \num{1.93e-06}$\pm$ \num{7.31e-13} 
    \\ \hline
     \multirow{2}{*}{Lower}
    & mean 
    & \num{8.97e-05}$\pm$\num{4.40e-09}
    & \num{4.85e-06}$\pm$\num{3.88e-10}
    & \num{5.10e-07}$\pm$\num{3.13e-11}
    \\ \cline{2-5}
    & variance 
    & \num{4.74e-07 }$\pm$ \num{4.99e-14}
    & \num{2.24e-07}$\pm$ \num{2.52e-15} 
    & \num{1.96e-07}$\pm$ \num{6.30e-16} 
    \\ \hline
    \end{tabular}
\end{table*}

\subsection{GenNorm gradient compression performance}

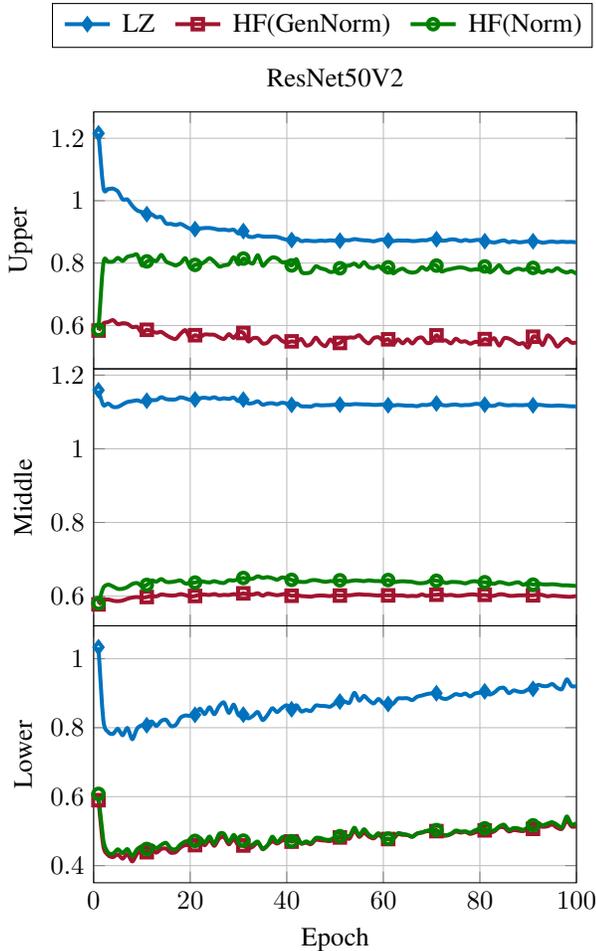
\begin{figure} 
    \centering
    \begin{tikzpicture}
    \definecolor{mycolor1}{rgb}{0.00000,0.44706,0.74118}%
    \definecolor{mycolor2}{rgb}{0.63529,0.07843,0.18431}%
    \definecolor{mycolor3}{rgb}{0.00000,0.49804,0.00000}%
    \begin{groupplot}[
        group style={
            group name=my plots,
            group size=1 by 3,
            xlabels at=edge bottom,
            xticklabels at=edge bottom,
            vertical sep=0pt
        },
        height=5cm,
        width=8cm,
        xmax=100,
        xmin=0,
        xlabel=Epoch,
        grid=both
    ]
    \nextgroupplot[title=ResNet50V2,
    ylabel=Upper]
        \coordinate (top) at (axis cs:1,\pgfkeysvalueof{/pgfplots/ymax});
        \addplot [draw=mycolor1, line width=1.5pt, mark=diamond, mark options={solid, mycolor1}, mark repeat={10}, smooth]
            table[x index=0, y index=1]{./Data/ResNet50V2_B1_C1_compression_rate.txt};
            \label{plots:plot1}
        \addplot [draw=mycolor2, line width=1.5pt, mark=square, mark options={solid, mycolor2}, mark repeat={10}, smooth]
            table[x index=0, y index=2]{./Data/ResNet50V2_B1_C1_compression_rate.txt};
            \label{plots:plot2}
        \addplot [draw=mycolor3, line width=1.5pt, mark=o, mark options={solid, mycolor3}, mark repeat={10}, smooth]
            table[x index=0, y index=3]{./Data/ResNet50V2_B1_C1_compression_rate.txt};
            \label{plots:plot3}
    \nextgroupplot[ylabel style={align=center},
    ylabel=Middle]
        \coordinate (top) at (axis cs:0,\pgfkeysvalueof{/pgfplots/ymax});
         \addplot [draw=mycolor1, line width=1.5pt, mark=diamond, mark options={solid, mycolor1}, mark repeat={10}, smooth]
            table[x index=0, y index=1]{./Data/ResNet50V2_B2_C1_compression_rate.txt};
        \addplot [draw=mycolor2, line width=1.5pt, mark=square, mark options={solid, mycolor2}, mark repeat={10}, smooth]
            table[x index=0, y index=2]{./Data/ResNet50V2_B2_C1_compression_rate.txt};
        \addplot [draw=mycolor3, line width=1.5pt, mark=o, mark options={solid, mycolor3}, mark repeat={10}, smooth]
            table[x index=0, y index=3]{./Data/ResNet50V2_B2_C1_compression_rate.txt};
    \nextgroupplot[ylabel style={align=center},
    ylabel=Lower]
        \addplot [draw=mycolor1, line width=1.5pt, mark=diamond, mark options={solid, mycolor1}, mark repeat={10}, smooth]
            table[x index=0, y index=1]{./Data/ResNet50V2_B4_C1_compression_rate.txt};
        \addplot [draw=mycolor2, line width=1.5pt, mark=square, mark options={solid, mycolor2}, mark repeat={10}, smooth]
            table[x index=0, y index=2]{./Data/ResNet50V2_B4_C1_compression_rate.txt};
        \addplot [draw=mycolor3, line width=1.5pt, mark=o, mark options={solid, mycolor3}, mark repeat={10}, smooth]
            table[x index=0, y index=3]{./Data/ResNet50V2_B4_C1_compression_rate.txt};
            
    \coordinate (bot) at (axis cs:1,\pgfkeysvalueof{/pgfplots/ymin});
    \end{groupplot}
    \path (top|-current bounding box.north)--
          coordinate(legendpos)
          (bot|-current bounding box.north);
    \matrix[
        matrix of nodes,
        anchor=south,
        draw,
        inner sep=0.2em,
        draw
      ]at([yshift=1ex,xshift=19ex]legendpos)
      {
        \ref{plots:plot1}& LZ &[3pt]
        \ref{plots:plot2}& HF(GenNorm) &[3pt]
        \ref{plots:plot3}& HF(Norm) \\};
\end{tikzpicture}
    \vspace{-0.3cm}
    \caption{Gradient compression ratio for upper, middle, and lower layers from the ResNet50V2.}
    \label{fig:gradient_compression_rate}
\end{figure}

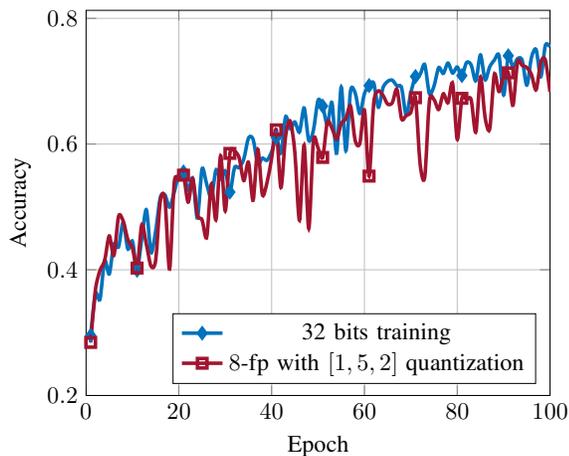
\begin{figure}
    \centering
	\begin{tikzpicture}[scale = 0.9]
    \definecolor{mycolor1}{rgb}{0.00000,0.44706,0.74118}%
    \definecolor{mycolor2}{rgb}{0.63529,0.07843,0.18431}%
    \definecolor{mycolor3}{rgb}{0.00000,0.49804,0.00000}%
    \begin{axis}[
    ymin=0.2,
    xmin=0,
    xmax=100,
    xlabel={Epoch},
    ylabel={Accuracy},
    legend pos=south east,
    grid=both
    ]
        \addplot [draw=mycolor1, line width=1.5pt, mark=diamond, mark options={solid, mycolor1}, mark repeat={10}, smooth]
            table[x index=0, y index=1]{./Data/Accuracy.txt};
            \addlegendentry{32 bits training}
        \addplot [draw=mycolor2, line width=1.5pt, mark=square, mark options={solid, mycolor2}, mark repeat={10}, smooth]
            table[x index=0, y index=2]{./Data/Accuracy.txt};
            \addlegendentry{$8$-fp with $[1,5,2]$ quantization}
    \end{axis}
\end{tikzpicture}
    \vspace{-0.3cm}
    \caption{Test accuracy of ResNet50V2.}
    \label{fig:accuracy}
\end{figure}

Although we are unable to substantiate the $\gennorm$ assumption for a large class of networks architectures and training datasets, we can argue that treating the DNN gradients as $\gennorm$ allows one to compress the gradients  more effectively, both in terms of compression rate and computational complexity.
In this section we consider the training performance for the model in Sec. \ref{sec:System Model}  with the rate in \eqref{eqn:Overhead}.
For this scenario, we wish to compare the compression performance of three compression schemes $h$:
\begin{enumerate}
    \item Compression using LZ78,
    \item Huffman coding  using $\gennorm$ modelling,
    \item Huffman coding using $\norm$ modelling.
\end{enumerate}
For 2) and 3) we use the quantized levels to compute the PMF of each bins from the CDF of fitted distribution.
%
In Fig. \ref{fig:gradient_compression_rate} we plot the compression performance for ResNet50V2  of the three schemes above.
We notice that the compression performance of the $\gennorm$ modelling provides an increase in performance at  very low computational cost.
As argued in Sec. \ref{sec:GenNorm modeling}, the gradient distribution approaches the $\norm$ distribution as the depth of the network increases: this can also be observed from the compression performance in the last panel in Fig. \ref{fig:gradient_compression_rate}.
In addition to smaller communication overhead, compression with Huffman coding also enjoys much lower complexity as compared to compression with LZ78 as the latter has to reconstruct the codebook on-the-fly.
Finally, in Fig.  \ref{fig:accuracy}, we plot the accuracy of the network trained with the $8$ bits quantized gradients as a function of the epoch number. 
We would like to emphasize that the loss in accuracy comes purely from the quantization and has nothing to do with data compression as lossless data compression is adopted.






\section{Conclusion}
In this paper, we have investigated the assumption that the DNN gradients can be well-modelled through a generalized normal distribution.
This observation has been used to implement gradient compression schemes for the rate-limited decentralized DNN training, that is the scenario in which a central DNN model is trained at remote users over  local datasets.
Once the DNN gradients have been obtained, the remote user quantizes these values and compresses them for transmission to the PS over a noiseless but rate limited communication channel.
A series of simulations have been conducted to validate that the gradient can be modelled as having $\gennorm$ distribution. 
Numerical evaluations have shown the effectiveness of this modelling in reducing the communication overhead between the parameter server and the remote users in the above scenario.

\bibliographystyle{IEEEtran}
\bibliography{ICC_zhong_jin }

\end{document}